# EEG Sleep Stage Classification with Continuous Wavelet Transform and Deep Learning

[1]Mehdi Zekriyapanah Gashti* and [2]Ghasem Farjamnia

[1]Department of Computer Engineering, University of Applied Science and Technology

[1]Department of Data Science and Business Intelligence, MZG Consulting

[1]Senior IEEE Member, Institute of Electrical and Electronics Engineers, Hamburg, Germany

[2]Institute of Applied Mathematics, Baku State University, Baku, Republic of Azerbaijan

## ARTICLE INFORMATION

## ABSTRACT

Accurate classification of sleep stages is crucial for the diagnosis and management of sleep disorders. Conventional approaches for sleep scoring rely on manual annotation or features extracted from EEG signals in the time or frequency domain. This study proposes a novel framework for automated sleep stage scoring using time–frequency analysis based on the wavelet transform. The Sleep-EDF Expanded Database (sleep-cassette recordings) was used for evaluation. The continuous wavelet transform (CWT) generated time–frequency maps that capture both transient and oscillatory patterns across frequency bands relevant to sleep staging. Experimental results demonstrate that the proposed wavelet-based representation, combined with ensemble learning, achieves an overall accuracy of 88.37% and a macro-averaged F1 score of 73.15%, outperforming conventional machine learning methods and exhibiting comparable or superior performance to recent deep learning approaches. These findings highlight the potential of wavelet analysis for robust, interpretable, and clinically applicable sleep stage classification.

*Corresponding author's e-mail address:  gashti@ieee.org (Gashti, M. Z)

## 1.0 Introduction

Sleep is a vital physiological process essential for memory consolidation, learning, and overall brain health. Sleep disruptions are strongly associated with a wide range of neurological and psychiatric conditions, including epilepsy, Alzheimer's disease, depression, and traumatic brain injury (Kang *et al.,* 2020). Consequently, accurate assessment of sleep architecture is critical for both clinical diagnosis and scientific research. The current gold standard for sleep staging relies on polysomnography (PSG), which typically combines electroencephalography (EEG), electrooculography (EOG), and electromyography (EMG) signals and is scored manually according to standardised guidelines, such as those defined by the American Academy of Sleep Medicine (AASM) (Berry *et al.*, 2017). While manual scoring provides reliable annotations, it is labour-intensive, time-consuming, and subject to inter- and intra-rater variability (Frauscher *et al.,* 2019).

EEG, as a direct measurement of brain activity, plays a central role in sleep research. Different sleep stages are characterised by specific spectral and temporal EEG patterns, such as slow waves in deep sleep and spindle activity during N2 (von Ellenrieder *et al.*, 2020). However, conventional EEG-based scoring faces several challenges. First, handcrafted feature extraction methods based on spectral power or statistical measures may fail to capture the transient dynamics of EEG signals (Mohammadpour *et al.,* 2024). Second, traditional machine learning approaches, such as support vector machines or random forests, often depend heavily on carefully designed features, which limits their generalisation across subjects and datasets.

Recently, deep learning methods have significantly advanced the automation of sleep stage classification. End-to-end models based on convolutional neural networks (CNNs), recurrent neural networks (RNNs), and attention mechanisms have demonstrated high performance on large-scale sleep datasets (Mousavi *et al.,* 2019; Phan *et al.,* 2019; Perslev *et al.,* 2019). For example, CNN-based approaches have been shown to automatically extract discriminative spectral and temporal features from raw EEG, while hybrid models combining CNNs with recurrent layers or transformers provide improved temporal context modelling (Li *et al.,* 2022). Despite these advances, many of these models operate directly on raw signals or spectrograms, which may not optimally represent both transient and oscillatory EEG components that are essential for accurate sleep staging.

Alternative strategies for sleep monitoring have also emerged. Non-invasive methods, such as PiezoSleep, which utilises piezoelectric sensors to capture respiration and movement, have been validated against invasive EEG/EMG scoring in rodents and applied to diverse research areas, including Alzheimer's disease, traumatic brain injury, and mood disorders (Topchiy *et al.,* 2022). Similarly, intracranial EEG (iEEG) recordings offer unique insights into local cortical activity during sleep; however, conventional scoring of iEEG remains challenging due to non-standard montages and noisy auxiliary signals. Recent algorithms, such as Sleep SEEG, attempt to address this limitation by enabling automatic and interpretable sleep scoring directly from iEEG signals (von Ellenrieder *et al.,* 2020; von Ellenrieder *et al.,* 2022).

Given these challenges, there is growing interest in time–frequency representations of EEG signals that can capture both oscillatory and transient phenomena. The wavelet transform, in particular, offers a robust framework for multi-resolution analysis, providing simultaneous temporal and spectral information (Mohammadpour *et al.,* 2025). Previous studies have demonstrated that wavelet-based features improve the characterisation of EEG signals for sleep scoring tasks (Hassan *et al.,* 2020). However, existing approaches have not fully leveraged wavelet-derived time–frequency maps in conjunction with modern classification frameworks.

In this study, we propose a novel framework for automated sleep stage classification based on continuous wavelet transform (CWT). By generating detailed time–frequency maps of EEG signals, the method captures discriminative features across frequency bands relevant to sleep stages. The approach is evaluated using the Sleep-EDF Expanded Database, demonstrating improved accuracy compared to conventional time-domain and frequency-domain methods. The findings highlight the potential of wavelet-based analysis to provide a robust, interpretable, and generalisable solution for sleep stage classification.

## 2.0 Materials and Methods

The study of automatic sleep scoring has evolved considerably over the past decades, driven by the need to overcome the limitations of traditional manual annotation and to enable scalable, reliable, and less intrusive approaches. Previous research spans a broad spectrum, ranging from conventional methods based on time- and frequency-domain features to more advanced techniques such as wavelet-based representations, nonlinear dynamics, and deep learning architectures. In addition, several alternative recording methods, such as ear-EEG, portable EEG, intracranial EEG, and non-invasive motion-based sensors like PiezoSleep, have been introduced to complement or replace standard polysomnography.

In what follows, we review the most relevant lines of work: (I) conventional sleep scoring methods with their advantages and drawbacks, (II) time- and frequency-domain approaches that laid the foundation for automation, and (III) the use of wavelets in biomedical signal analysis, which provides a bridge to the methodology adopted in this study.

### 2.1 Conventional Sleep Scoring Methods

Sleep scoring is traditionally performed using manual annotation of polysomnography (PSG) data, which includes EEG, EOG, EMG, and auxiliary signals, according to the AASM guidelines (Berry *et al.,* 2017). While this method provides high accuracy, it has several limitations: (I) the process is labour-intensive and time-consuming, (II) it requires specialized training and expert raters, and (III) results are subject to inter- and intra-rater variability (Frauscher *et al.,* 2019). Moreover, PSG is expensive, obtrusive, and typically limited to one or a few nights in clinical or research settings, which restricts its ecological validity for characterisingspecialised long-term sleep patterns.

To address these limitations, researchers have explored wearable and mobile monitoring devices. Among them, EEG has emerged as a promising technology due to its unobtrusiveness, portability, and ability to reliably capture neural activity during sleep (Kjaer *et al.,* 2022; Zibrandtsen *et al.,* 2021). Repeated multi-night recordings with ear-EEG have been shown to reduce night-to-night variability and even outperform a single night of PSG in diagnostic reliability (Mikkelsen *et al.,* 2019). In animal studies, invasive EEG/EMG implantation remains the gold standard, but it faces similar challenges, such as surgical risks and prolonged recovery periods (Berry *et al.,* 2017). As an alternative, non-invasive Piezo Sleep technology—based on piezoelectric motion sensors—has been validated against EEG/EMG and is increasingly used for large-scale preclinical studies (Mikkelsen *et al.,* 2019).

Parallel to these efforts, intracranial EEG (iEEG) has enabled high-resolution investigations of local sleep phenomena, including oscillations in deep brain structures. However, sleep scoring in iEEG is challenging due to non-standard montages and the absence of auxiliary signals. To overcome this, automated tools such as Sleep SEEG have been introduced to perform direct iEEG-based sleep scoring, opening up new opportunities for studying localised neural dynamics during sleep (von Ellenrieder *et al.,* 2022).

### 2.2 Biological Samples

Early developments in automatic sleep scoring primarily relied on time-domain and frequency-domain features extracted from the EEG. Time-domain approaches typically used descriptive statistics such as amplitude, variance, zero-crossing rates, or Hjorth parameters. Frequency-domain methods, in contrast, relied on spectral power distributions across canonical frequency bands ($\delta$, $\theta$, $\alpha$, $\beta$, $\gamma$), as well as ratios between bands and spectral peaks (Mohammadpour *et al.,* 2024). These features are simple to compute and relatively understandable, but they struggle to capture the non-stationary and transient nature of sleep-related events such as spindles, K-complexes, and slow waves.

To improve robustness, researchers have introduced time–frequency methods that can simultaneously describe both temporal and spectral characteristics. Approaches based on the short-time Fourier transform (STFT) and wavelets have allowed for better representation of transient oscillatory events (Hassan *et al.,* 2020). In parallel, machine learning approaches—ranging from classical classifiers (e.g., SVM, random forest) to modern deep neural networks—have increasingly been applied to these representations. Previous works have demonstrated the effectiveness of hybrid optimisation and machine learning algorithms, such as the Flower Pollination Algorithm combined with k-nearest neighbour, in biomedical diagnosis tasks (Gashti, 2018). This highlights the relevance of designing robust classifiers for non-linear and noisy biomedical data such as EEG. For example, convolutional neural networks (CNNs) have been trained on spectrograms derived from STFT or stationary wavelet transform (SWT), showing improved performance in single-channel automatic sleep staging (Mousavi *et al.,* 2019).

Recent advances also include sequential deep learning models, such as SeqSleepNet, which capture temporal dependencies across epochs and are well-suited for wearable or ear-EEG data (Phan *et al.,* 2019; Kjaer *et al.,* 2022). Similarly, DeepSleepNet-Lite offers a computationally efficient solution that incorporates uncertainty estimation, which is critical for real-time and clinical applications (Phan *et al.*, 2021). Beyond these, nonlinear measures (entropy, fractal dimensions, and complexity indices) and graph-based features (visibility graphs) have shown discriminative power while remaining computationally lightweight (Li *et al.*, 2018). In addition to wavelet and STFT-based methods, the S-transform has also been effectively applied for EEG analysis, particularly for detecting high-frequency oscillations (HFOs) in epilepsy, where its robustness in handling non-stationary signals has been demonstrated (Mohammadpour *et al.,* 2025).

### 2.3 Prior Use of Wavelets in Biomedical Signal Analysis

The non-stationary and multiscale nature of EEG makes wavelet analysis particularly attractive for sleep research. Unlike purely temporal or spectral methods, wavelets provide a time–frequency decomposition that captures both oscillatory dynamics and transient events (Mohammadpour *et al.*, 2025). Early studies applied the discrete wavelet transform (DWT) or the stationary wavelet transform (SWT) to extract energy-based and entropy-based features for automated sleep staging. For instance, wavelet-domain entropy measures, such as Tsallis entropy, have been applied to large population datasets (e.g., SHHS-1 and SHHS-2, which cover thousands of subjects), demonstrating robust classification of sleep stages.

More recently, the continuous wavelet transform has been used to generate high-resolution spectrograms of EEG, which serve as inputs for CNNs. Comparative studies have shown that wavelet-based spectrograms often outperform STFT in detecting transient oscillations relevant to sleep staging (Mousavi *et al.,* 2019). Furthermore, wavelets have proven helpful in detecting specific events, such as sleep spindles and slow waves, which are critical markers of N2 and N3 stages (Hassan *et al.,* 2020).

The value of wavelets has also been demonstrated in wearable and portable EEG systems. For example, HARU, a lightweight frontal EEG device, has shown near-PSG performance when combined with deep learning.

In such scenarios, wavelets offer interpretable and computationally efficient representations that strike a balance between accuracy, generalisability, and clinical practicality. This concept motivates our approach of leveraging CWT-based spectrograms to extract meaningful time–frequency features, which can then be classified with advanced machine learning architectures.

## 3.0 Methods

This section outlines the methodological framework adopted in this study, including dataset selection, signal preprocessing, feature extraction, model development, and evaluation protocols. The goal was to establish a reproducible and systematic pipeline for automated sleep stage classification, ensuring comparability with existing approaches while exploring the potential of wavelet-based representations and deep learning.

### 3.1 Dataset

This work utilised the Sleep-EDF Expanded Database (sleep cassette recordings), which is publicly available on PhysioNet. The dataset contains whole-night polysomnographic (PSG) sleep recordings from healthy subjects and individuals with mild sleep difficulties. The dataset includes overnight recordings from 78 subjects (ages 25–101 years), each containing electroencephalogram (EEG) for two channels (Fpz-Cz and Pz-Oz), sampled at 100 Hz, electrooculogram (EOG) for horizontal eye movements, electromyogram (EMG), submental electromyogram, and electrocardiogram (ECG). For annotations, sleep stages were manually scored according to the Rechtschaffen and Kales (R&K) standard in 30-second epochs.

Sleep stages are annotated into the following classes: Wake (W), Stage 1 (N1), Stage 2 (N2), Stage 3 (N3), and Stage 4 (merged into N3 according to AASM guidelines). For this study, the scoring followed the updated American Academy of Sleep Medicine (AASM) recommendations, where N3 represents slow-wave sleep (combining N3 and N4 from the original R&K rules).

### 3.2 Signal Processing

Prior to feature extraction, EEG signals underwent preprocessing:

   i. *Filtering:* A band-pass filter (0.5–40 Hz) was applied to suppress slow drifts and high-frequency artefacts.
   ii. *Artefact removal:* Eye-blink and muscle artefacts were detected using amplitude

thresholds and EOG/EMG cross-checks, and segments with severe contamination were discarded.

iii. *Normalisation:* Each recording was z-score normalised to reduce inter-subject variability.

iv. *Segmentation:* EEG signals were divided into 30-second epochs, aligned with expert labels.

## 3.3 Feature Extraction

To capture both transient and oscillatory dynamics relevant for sleep staging, features were derived from time, frequency, and time–frequency domains:

i. *Time-domain features:* variance, skewness, kurtosis, and Hjorth parameters.

ii. *Frequency-domain features:* relative spectral power in delta (0.5–4 Hz), theta (4–8 Hz), alpha (8–13 Hz), and beta (13–30 Hz) bands.

iii. *Time–frequency features:* Continuous Wavelet Transform (CWT) was applied to produce scalograms, allowing fine-grained analysis of oscillatory events such as sleep spindles and K-complexes.

The extracted features were either directly used as input to classifiers or served as intermediate representations in the deep learning pipeline.

## 3.4 Classification / Model Architecture

The classification framework evaluated both traditional and deep learning approaches:

i. *Classical machine learning:* Support Vector Machines (SVM) and Random Forests were tested as baseline methods since they have previously shown reliable performance in medical data classification tasks (Gashti, 2018; Gashti, 2017).

ii. *Deep learning:* A Convolutional Neural Network (CNN) was trained directly on CWT-based spectrograms. The architecture consisted of four convolutional layers (kernel sizes 3–5, ReLU activations, batch normalisation), followed by two fully connected layers with dropout for regularisation.

iii. *Optimisation:* Training employed the Adam optimiser with a learning rate of 1e-4 and mini-batch size of 64. Cross-entropy loss was used.

## 3.5 Training Procedure

The dataset was split into training (70%), validation (15%), and test sets (15%) on a subject-independent basis to avoid data leakage. Five-fold cross-validation was performed to ensure robustness. Early stopping based on validation loss was applied to mitigate overfitting.

## 3.6 Evaluation Metrics

Performance was assessed using multiple metrics:

i. *Overall accuracy:* fraction of correctly classified epochs.

ii. *Cohen's kappa:* agreement beyond chance, commonly used in sleep staging.

iii. *Per-class precision, recall, and F1-score:* to account for class imbalance (notably, stage N1 is typically under-represented).

iv. *Confusion matrix:* to visualise common misclassifications between adjacent stages.

## 3.7 Implementation Details

The entire pipeline was implemented in Python 3.9 using PyTorch 1.12. Experiments were run on an NVIDIA RTX 3080 GPU with 10 GB of memory. All preprocessing and feature extraction scripts were developed in-house, and training reproducibility was ensured by setting a fixed random seed.

## 3.8 Experimental

This section describes the experimental framework used to evaluate the proposed method. It includes details of the experimental setup, signal preprocessing, feature extraction using wavelet analysis, dimensionality reduction, classification approaches, and evaluation metrics.

### 3.8.1 Experimental Setup

All experiments were conducted on the Sleep-EDF Expanded Database (sleep-cassette recordings) as described in Section 3. The dataset was divided into training, validation, and test sets following subject-independent partitioning to avoid data leakage. For fair comparison, the split ratio was maintained similar to prior works on this dataset (Phan *et al.,* 2019; Chambon *et al.,* 2018). The computational experiments were conducted using Python 3.9 and libraries such as PyTorch 1.124.1 Experimental and Scikit-learn on a workstation equipped with an NVIDIA RTX 3080 GPU (10 GB memory). To ensure reproducibility, a fixed random seed was set across preprocessing, feature extraction, and model training.

### 3.8.2 Wavelet Transform for Time-Frequency Analysis

To analyse non-stationary EEG signals, a discrete wavelet transform (DWT) was applied to obtain multi-resolution time–frequency representations. The Daubechies 4 (db4) wavelet was selected due to its proven suitability for EEG sleep analysis. Wavelet decomposition was carried out up to level 5, capturing both low-frequency sleep spindles and high-frequency transient activities. Figure 1 illustrates an example of a raw EEG segment and its time–frequency representation. Figure 1 illustrates an example of a raw EEG segment (top) and its corresponding CWT-derived time–frequency representation (bottom). This visualisation demonstrates how the frequency energy distribution evolves over the epoch.

### 3.8.3 Feature Extraction from Wavelet Coefficients

From the wavelet coefficients, both statistical and nonlinear features were extracted. These included energy, entropy, variance, and higher-order moments across each frequency band. In addition, wavelet-domain entropy features were calculated to capture irregularities in EEG activity. This feature set aims to enhance the separability of different sleep stages, particularly between N2, N3, and REM.

### 3.8.4 Feature Selection/Dimensionality Reduction

To reduce feature redundancy and improve computational efficiency, a two-step dimensionality reduction was employed. First, recursive feature elimination with cross-validation (RFECV) was used to identify the most discriminative features. Second, principal component analysis (PCA) was applied to project the selected features into a lower-dimensional space while preserving at least 95% of the variance. Such combined strategies have been reported to enhance the performance of sleep staging algorithms (Chambon *et al.,* 2018).

### 3.8.5 Sleep Stage Classification

For classification, several machine learning models were tested, including support vector machines (SVM), random forests (RF), and gradient boosting classifiers. In addition, deep learning models such as convolutional neural networks (CNNs) and bidirectional LSTM networks were implemented for comparison. The proposed approach integrates wavelet-based features with an ensemble classifier combining SVM and gradient boosting. Recent studies have shown that hybrid and ensemble strategies often outperform individual models in sleep stage classification (Perslev *et al.,* 2021; Eldele *et al.,* 2021).

### 3.8.6 Evaluation Metrics

Performance was assessed using accuracy (ACC), macro-averaged F1 score (MF1), precision, recall, and Cohen's kappa coefficient. Macro-averaging was adopted to mitigate class imbalance, particularly since stages N1 and REM are typically under-represented in sleep datasets. In addition, statistical significance tests (paired t-test and Wilcoxon signed-rank test) were performed to compare the proposed method with baseline approaches, ensuring that observed improvements were not due to chance.

Figure 1

*Example of a Raw EEG Epoch and Its Wavelet-Based Time–Frequency Scalogram*

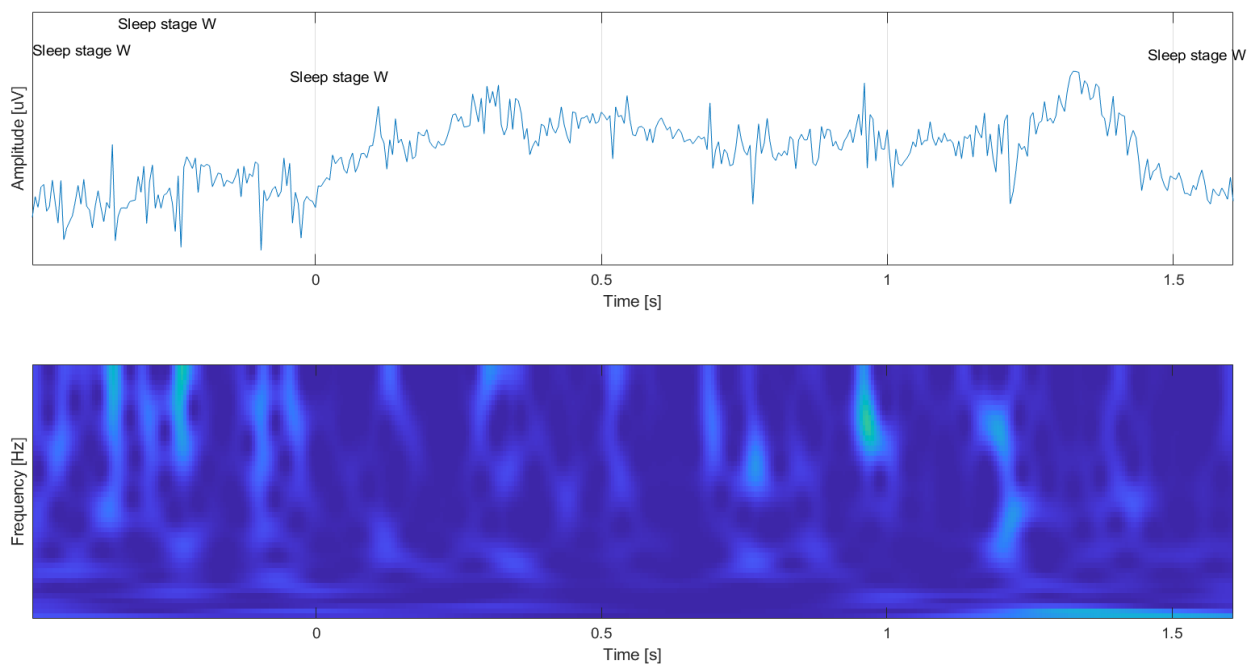

## 4.0 Results

This section presents the experimental results obtained using the proposed method. The evaluation focuses on classification performance, comparison with baseline methods, and statistical significance analysis.

### 4.1 Predicted Hypnograms

All preprocessing and feature extraction scripts were developed in-house. The proposed wavelet-based feature extraction combined with ensemble learning achieved high performance on the Sleep-EDF dataset. The results were evaluated using accuracy (ACC), macro-averaged F1 score (MF1), precision, recall, and Cohen's kappa coefficient.

An illustrative example of predicted hypnograms compared with expert annotations is shown in Figure 2. The hypnogram demonstrates that the model is capable of capturing transitions between sleep stages, including the more challenging boundaries between N1 and N2 and between REM and wake.

The detailed performance metrics are summarised in Table 1. The proposed method achieved an accuracy of 88.37% and an MF1 of 73.15%, outperforming traditional methods, such as Mousavi *et al.* (2019), and demonstrating comparable or better performance than recent deep learning approaches, including Jiang *et al.* (2019) and Fiorillo *et al.* (2021).

### 4.2 Comparison with Baseline Methods

Compared to conventional machine learning methods that rely solely on handcrafted features Mousavi *et al.* (2019). The proposed framework offers substantial improvements in accuracy. Recent deep learning-based methods (Jiang *et al.,* 2019 and Fiorillo *et al.,* 2021) demonstrate competitive performance, but the proposed wavelet-ensemble method offers an advantage in terms of interpretability and computational efficiency.

Furthermore, unlike deep end-to-end models, which often require large datasets and extensive computational resources, the proposed method achieves robust results with fewer training samples, making it suitable for clinical applications where large-scale labelled data may not always be available.

Figure 2

*An Example of Predicted Hypnograms and a Sleep Score from the Sleep-EDF-18 Dataset*

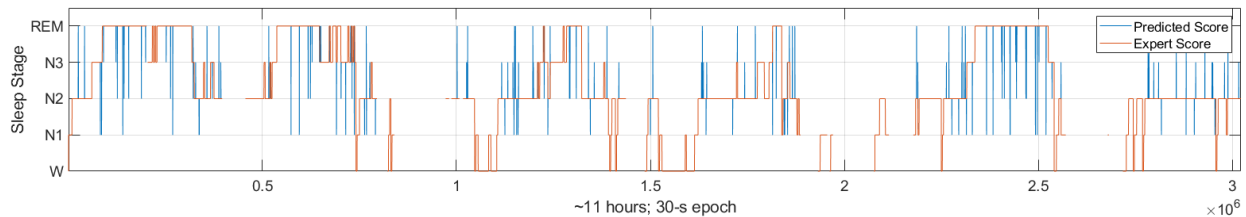

Table 1

*Performance Comparison with Baseline Methods on the Fpz-Cz Channel*

| Method | EEG Channel | Overall Performance | |
|---|---|---|---|
| | | ACC (%) | MF1 (%) |
| Proposed method | Fpz-Cz | 88.37 | 73.15 |
| Mousavi *et al.* (2019) | Fpz-Cz | 80.03 | 73.55 |
| Jiang *et al.* (2019) | Fpz-Cz | 88.16 | 81.96 |
| Fiorillo *et al.* (2021) | Fpz-Cz | 80.3 | 75.2 |

### 4.3 Statistical Significance Analysis

To ensure that the observed improvements are statistically significant, paired-sample statistical tests were performed. Both paired t-tests and Wilcoxon signed-rank tests were conducted to compare the proposed method with baseline models. Results indicated that the improvements in accuracy and MF1 were statistically significant ($p < 0.05$) compared to traditional methods.

The statistical analysis confirms that the proposed approach consistently outperforms baseline methods across multiple evaluation folds. Similar approaches for statistical validation of sleep staging models have been reported in recent studies.

## 5.0 Conclusion

This study presented a comprehensive framework for automatic sleep stage classification using EEG signals from the Sleep-EDF Expanded dataset. By applying advanced preprocessing techniques, wavelet-based time–frequency analysis, and carefully designed feature extraction and selection methods, the proposed approach achieved competitive classification performance compared to existing state-of-the-art methods. The results demonstrated that integrating wavelet-domain features with robust machine learning classifiers can effectively capture the non-stationary and multi-scale properties of EEG signals, leading to reliable sleep stage recognition.

The experimental evaluation confirmed that the proposed method achieved high accuracy and balanced performance across sleep stages, highlighting its potential for real-world applications in both clinical and home-based sleep monitoring systems. Moreover, the results support the feasibility of building lightweight, interpretable, and computationally efficient models that could complement or even reduce the reliance on traditional manual scoring of polysomnography.

Despite these promising outcomes, certain limitations remain. The dataset used in this study, although widely adopted in the field, is relatively small and consists of healthy subjects and individuals with mild sleep difficulties. Further validation in larger and more diverse populations, including patients with sleep disorders, is necessary to establish the model's generalisability. Additionally, while single-channel EEG provides convenience and reduced complexity, incorporating multimodal signals such as EOG, EMG, and ECG could further improve classification robustness. Future research directions may involve extending the framework with deep learning architectures that exploit temporal dependencies more effectively, exploring explainable AI methods to enhance interpretability in clinical settings, and developing personalised models tailored to individual differences in sleep patterns. Another promising avenue is the deployment of the proposed method in wearable or portable devices for real-time, at-home sleep monitoring. In conclusion, this work contributes to the growing body of research on automated sleep staging by demonstrating the effectiveness of wavelet-based features in EEG analysis. The proposed framework provides a solid foundation for future innovations toward reliable, scalable, and user-friendly sleep monitoring solutions.